\documentclass[10pt,journal,compsoc]{IEEEtran}
\renewcommand{\baselinestretch}{0.98}
\ifCLASSOPTIONcompsoc
  \usepackage[nocompress]{cite}
\else
  \usepackage{cite}
\fi
\usepackage{ragged2e}
\usepackage{graphicx}
\usepackage{comment}
\usepackage{amsmath,amssymb} 

\usepackage[switch]{lineno}

\usepackage{multirow}
\usepackage{multicol}
\usepackage{floatrow}
\usepackage[noend]{algpseudocode}
\usepackage{algorithmicx,algorithm}
\usepackage{bbding}
\usepackage{bm}
\usepackage{stfloats}
\usepackage{soul,tikz,color,xcolor,hyperref}
\usepackage{subfigure}
\usepackage{booktabs}
\usepackage{makecell}
\usepackage{xspace}

\definecolor{lime}{HTML}{A6CE39}
\DeclareRobustCommand{\orcidicon}{%
    \begin{tikzpicture}
    \draw[lime, fill=lime] (0,0) 
    circle [radius=0.16] 
    node[white] {{\fontfamily{qag}\selectfont \tiny ID}};    \draw[white, fill=white] (-0.0625,0.095) 
    circle [radius=0.007];    \end{tikzpicture}
    \hspace{-2mm}}
\foreach \x in {A, ..., Z}{%
    \expandafter\xdef\csname orcid\x\endcsname{\noexpand\href{https://orcid.org/\csname orcidauthor\x\endcsname}{\noexpand\orcidicon}}
    }

\begin{document}
%
\title{SkinGPT-X: A Self-Evolving Collaborative Multi-Agent System for Transparent and Trustworthy Dermatological Diagnosis}
%
%
%
%

\author{Zhangtianyi~Chen$^{1,\dagger}$, Yuhao Shen$^{1,\dagger}$, Florensia Widjaja$^{1,\dagger}$, Yan Xu$^{2}$, Liyuan Sun$^{3}$, Zijian Wang$^{1}$, Hongyi Chen$^{1}$, Wufei Dai$^{1}$, Juexiao~Zhou$^{1,*}$
\thanks{
$^1$School of Data Science, The Chinese University of Hong Kong, Shenzhen\\
$^2$Department of Dermatology, Tianjin Institute of Integrative Dermatology, Tianjin Academy of Traditional Chinese Medicine Affiliated Hospital, Tianjin 300120, China\\
$^3$Department of Dermatology, Beijing AnZhen Hospital, Capital Medical University, Beijing 100029, China..\\
$^\dagger$These authors contributed equally.\\
$^*$Corresponding author. e-mail: juexiao.zhou@gmail.com\\
}}
%
%

\markboth{}%
%
\\
\IEEEtitleabstractindextext{%
\begin{abstract}
\justifying
While recent advancements in Large Language Models have significantly advanced dermatological diagnosis, monolithic LLMs frequently struggle with fine-grained, large-scale multi-class diagnostic tasks and rare skin disease diagnosis owing to training data sparsity, while also lacking the interpretability and traceability essential for clinical reasoning. Although multi-agent systems can offer more transparent and explainable diagnostics, existing frameworks are primarily concentrated on Visual Question Answering and conversational tasks, and their heavy reliance on static knowledge bases restricts adaptability in complex real-world clinical settings. Here, \textbf{we present SkinGPT-X, a multimodal collaborative multi-agent system for dermatological diagnosis integrated with a self-evolving dermatological memory mechanism.} By simulating the diagnostic workflow of dermatologists and enabling continuous memory evolution, SkinGPT-X delivers transparent and trustworthy diagnostics for the management of complex and rare dermatological cases. To validate the robustness of SkinGPT-X, we design a three-tier comparative experiment.
First, we benchmark SkinGPT-X against four state-of-the-art LLMs across four public datasets, demonstrating its state-of-the-art performance with a +9.6\% accuracy improvement on DDI31 and +13\% weighted F1 gain on Dermnet over the state-of-the-art model. Second, we construct a large-scale multi-class dataset covering 498 distinct dermatological categories to evaluate its fine-grained classification capabilities. Finally, we curate the rare skin disease dataset, the first benchmark to address the scarcity of clinical rare skin diseases which contains 564 clinical samples with eight rare dermatological diseases. On this dataset, SkinGPT-X achieves a +9.8\% accuracy improvement, a +7.1\% weighted F1 improvement, a +10\% Cohen's Kappa improvement. The self-evolving agent memory enables the continuous accumulation of historical cases and the iterative evolution of diagnostic guidelines, which significantly enhances the system's reasoning depth across an expanding range of skin diseases.
In summary, SkinGPT-X represents the first multimodal collaborative multi-agent-driven dermatological diagnosis system empowered by a self-evolving memory mechanism, enabling transparent and trustworthy diagnosis, particularly for rare skin diseases affected by insufficient training data.
\end{abstract}

\begin{IEEEkeywords}
Dermatology, Multi-agent system, Large language model
\end{IEEEkeywords}}

\maketitle

\IEEEdisplaynontitleabstractindextext

%
\IEEEpeerreviewmaketitle

\section{Introduction}
\label{sec:intro}

Skin diseases exert a substantial global health burden, profoundly impacting not only physical well-being but also the psychosocial quality of life for millions \cite{hay2014global, salari2024global}. Recent updates from the Global Burden of Disease study further underscore that skin and subcutaneous diseases remain leading causes of non-fatal disability worldwide \cite{gorouhi2024global}. Despite this growing recognition of skin health as a pillar of overall well-being, the accessibility of specialized care remains severely constrained by a systemic shortage of licensed practitioners\cite{seth2017global}. Consequently, the diagnostic benefits for most minor skin ailments are disproportionately low compared to the time and effort patients expend to visit these major institutions\cite{emanuel2020measuring,warshaw2011teledermatology}. As the result, teledermatology becomes more and more popular in order to expand the range of services available to medical professionals\cite{lee2018teledermatology, tuckson2017advanced}. Yet, while teledermatology addresses the issue of geographical distance, it does little to alleviate the absolute scarcity of expert labor\cite{whited2001teledermatology}. In fact, the surge in digital consultations often overwhelms the existing clinical workforce\cite{moy2021virtual}. This bottleneck has necessitated the integration of automated diagnostic tools\cite{thirunavukarasu2023large} to facilitate triage and preliminary screening in dermatology.

Artificial Intelligence (AI) is widely regarded as having immense potential to augment human expertise in specific medical domains\cite{esteva2017dermatologist}. Its application in dermatology holds profound significance in reducing healthcare costs and enhancing diagnostic efficiency\cite{topol2019high, gomolin2020artificial, chan2020machine}. The breakthrough in Deep Learning (DL) has witnessed a paradigm shift, moving from traditional visual inspections to automated, high-precision diagnostic systems\cite{noronha2023deep}. Over the past decades, diverse DL architectures have been deployed across various dermatological domains. These include standard Convolutional Neural Networks (CNNs) such as VGG-16/19 \cite{vgg_app}, ResNet-50 \cite{resnet_app}, and Inception-V3 \cite{inception_app} for robust feature extraction, as well as lightweight models like MobileNet V2 \cite{mobilenet_app} and Xception \cite{xception_app} designed for mobile-based point-of-care diagnostics.
These DL-based methods have been proposed for various dermatological domains like skin disease classification \cite{bandyopadhyay2022, jha2022evolutionary, han2018, erkol2005, mazhar2023, lembhe2023, venugopal2023}, skin cancer diagnosis \cite{jha2022hybrid, haenssle2018, jaisakthi2023, zafar2023}, melanoma detection \cite{gilmore2010, marks2000, melarkode2023, gilani2023, priyadharshini2023}, and analysis for different types of psoriasis \cite{kadampur2020}.  Beyond classification, advanced frameworks involving Generative Adversarial Networks (GANs)\cite{gajera2023comprehensive} for data augmentation and hybrid deep learning approaches for skin lesion segmentation\cite{thapar2022retracted} have significantly improved the detection of malignant conditions. Despite achieving expert-level accuracy in controlled datasets, recent studies highlight critical ongoing challenges such as the need for cross-population generalization \cite{pacheco2020}, handling class imbalance in rare skin diseases \cite{tahir2023,jabbour2023deep}, and the imperative for explainable AI (XAI) to foster clinical trust \cite{melarkode2023, zafar2023}.

Recently, Large Language Models (LLMs) have emerged as a transformative force in artificial intelligence. Compared to traditional DL models, LLMs exhibit superior multimodal integration capabilities\cite{moor2023foundation, achiam2023gpt, singhal2023large, ji2023domain}, which can capture long-range dependencies between disparate data modalities, including clinical text, dermoscopic images, and omics data. 
The application of LLMs in dermatology extends beyond conventional skin cancer diagnosis \cite{de2020, cirone2024, liu2024, shifai2024, laohawetwanit2024} and clinical condition classification \cite{rundle2024, sievert2024, omiye2024}. These models also demonstrate significant potential to augment traditional healthcare workflows by providing medication guidance \cite{chen2024} and addressing common patient inquiries with high accessibility \cite{breneman2024, robinson2024, lauck2024}. Furthermore, LLMs offer a promising solution for bridging the communication gap between clinicians and patients by translating intricate medical terminology into easily comprehensible language \cite{el_saadawi2007, zhang2024, liopyris2022}. This enables a more holistic understanding of complex pathologies \cite{huang2023visual, tu2024towards} and facilitates precision medicine by tailoring treatment schedules \cite{lewandowski2025systemic}. Furthermore, LLMs serve as pivotal tools for patient education and physician-patient interaction by simplifying complex medical reports into patient-friendly language \cite{zhou2023skingpt4,zhang2023huatuogpt,shen2026trustworthyfairskingptr1democratizing}. Although LLMs have made significant progress, there are still two main limitations when applying them to clinical diagnosis. \textbf{1) Bottlenecks in High-Cardinality and Rare Disease Tasks:} The static nature of fine-tuned knowledge bases proves inadequate for large-scale diagnostic tasks involving an extensive label space and the identification of rare diseases. These models often fail to generalize across such a vast and fine-grained distribution of pathologies, falling short of the performance required for real-world clinical deployment\cite{greco2023limitations}. \textbf{2) Insufficient Traceability of Standalone Models:} While standalone LLMs can provide linguistic explanations, their descriptive granularity often falls short of rigorous clinical standards. Such models lack the necessary diagnostic traceability and evidence-based reasoning required for high-stakes medical accountability\cite{ghassemi2021false}.

Multi-Agent System represents a promising paradigm for overcoming these limitations by establishing a traceable chain of evidence through collaborative intelligence \cite{wang2023unleashing, chan2023chateval}. By distributing complex reasoning across a network of specialized agents where each assumes distinct roles such as clinical analysis, evidence retrieval, or cross-verification, multi-agent system simulates the consultative process of a multi-disciplinary medical team \cite{li2023camel, tang2023medagents}. This synergetic workflow allows the system to decompose intricate diagnostic tasks into discrete, cross-validated stages \cite{khot2022decomposed}, effectively transforming monolithic LLM outputs into structured, verifiable, and evidence-based diagnostic pipelines \cite{wang2023unleashing}. Currently, the use of multi-agent systems in healthcare is still in its infancy, largely limited to text-based conversations \cite{tang2023medagents}. However, in real-world settings like teledermatology, the ability to process multi-modal data is of paramount importance for diagnostic accuracy\cite{han2020deep}. Furthermore, Retrieval-Augmented Generation (RAG) is extensively employed to equip these agents with specialized medical knowledge \cite{zakka2024aloha, soong2024improving}. Despite these efforts, static knowledge bases typically encapsulate only standardized standards from textbooks, which often prove inadequate when confronted with atypical clinical manifestations or rare variant presentations in complex real-world scenarios \cite{jabbour2023deep}.
\begin{figure*}[p]
    \centering
    \vspace{-10mm}  
    \includegraphics[width=0.93\linewidth]{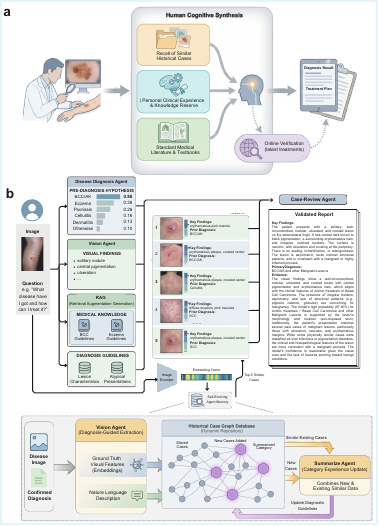} 
    \vspace{-4mm}  
    \caption{\textbf{a, The process of clinicians formulating a comprehensive Diagnosis Report}: Clinicians formulate a Diagnosis Report by integrating three core pillars of medical intelligence: Personal Clinical Experience, Standard Medical Literature, and the Recall of Similar Historical Cases. This holistic synthesis, further augmented by Online Verification of the latest treatments, enables a validated Treatment Plan. \textbf{b, The Architecture of the SkinGPT-X System}: Upon receiving a disease image and a query, the Vision Agent extracts fine-grained Visual Findings, while the Diagnosis Diagnosis Agent performs an initial Pre-diagnosis Hypotheses. This process is anchored by the RAG, which retrieves Local Medical Knowledge to ensure evidence-based reasoning. To incorporate empirical experience, the system utilizes the self-evolving agent memory, where the Top-5 Similar Cases and Diagnostic Guidelines are retrieved from the Historical Case Graph Database. The visual features, retrieved guidelines, and historical precedents are synthesized by the Case-Review Agent, which conducts a rigorous cross-reference to produce a validated Report. The Summarize Agent ensures self-evolving agent memory by integrating new confirmed cases into the Dynamic Repository, iteratively distill the diagnostic guidelines without retraining.}
    \label{fig:skingpt_framework}
\end{figure*}

To address these challenges, we propose a novel \textbf{multi-agent system architecture named SkinGPT-X}, inspired by authentic clinical diagnostic workflows. Our system harmonizes textbook-based expertise, empirical experience, and visual interpretation into a collaborative framework (Fig. \ref{fig:skingpt_framework}b). First, we decouple the pre-diagnostic inference from visual feature findings extraction by employing two specialized agents: one dedicated to extracting visual findings and the other to generating initial diagnostic hypothesis. To ensure foundational clinical accuracy, we employ a RAG module to anchor the initial reasoning process in evidence-based dermatological common sense\cite{zakka2024aloha}. Second, we propose a Self-\textbf{Evo}lving \textbf{Derma}tological Diagnostic \textbf{Mem}ory(\textbf{EvoDerma-Mem}) mechanism, which empowers agents to autonomously refine their internal knowledge bases without the need for parameter retraining. Unlike traditional systems that rely on static textbooks, SkinGPT-X iteratively synthesizes diagnostic experiences from encountered cases. As the volume and diversity of cases grow, the system generates comprehensive diagnostic guidelines that encapsulate both prototypical disease markers and rare clinical variants. This evolutionary approach ensures that the system continuously broadens its understanding of dermatological conditions, offering superior scalability and coverage compared to static knowledge base. During the inference phase, the system retrieves visually similar historical cases and corresponding guidelines from the agent memory. SkinGPT-X then cross-references these insights with the current visual features to conduct a rigorous review of the preliminary diagnosis. This architecture ensures that fine-grained disease subcategories and rare pathologies receive equitable attention. Furthermore, the self-generated agent memory acts as a corrective layer, enabling the system to detect subtle diagnostic oversights and evaluate the possibility of rare conditions, ultimately producing a more traceable, transparent, and trustworthy diagnostic report.

\section{Results}
\subsection{The overall design of SkinGPT-X}
SkinGPT-X operates as a self-evolving multi-agent system that bridges raw visual data with traceable clinical evidence as show in Fig. \ref{fig:skingpt_framework}. In the Diagnostic Phase, multiple agents collaborate to analyze patient images from diverse perspectives, including visual findings, preliminary diagnostic hypotheses and retrieval book-based medical knowledge. These insights are then audited by a Case-Review Agent, which is powered by a self-evolving agent memory to produce a final, validated report. The system’s Evolutionary Phase ensures continuous learning: new cases are archived as memory nodes, and upon reaching a knowledge density threshold, a Summarization Agent automatically synthesizes these historical cases into high-level clinical guidelines, emulating the expertise accumulation of a human dermatologist.

To rigorously evaluate the diagnostic efficacy, robustness, and interpretability of SkinGPT-X, we designed a three-tiered experimental framework ranging from common pathologies to complex rare diseases. In Section \ref{experiment1}, we conducted comprehensive benchmarking against three general medical LLMs and one fine-tuned foundation model (PanDerm\cite{yan2025multimodal}) across 4 public datasets (Dermnet\cite{dermnet}, HAM10000\cite{tschandl2018ham10000}, DDI\cite{daneshjou2022disparities}, and Fitzpatrick-17k\cite{groh2021evaluating}). Our findings indicate that SkinGPT-X not only excels in standard metrics like Accuracy (ACC) and Weighted F1 but also significantly outperforms baselines in Matthews Correlation Coefficient (MCC) and Cohen’s Kappa, suggesting its exceptional potential in addressing the fine-grained diagnosis task. Motivated by this observation, we sought to further explore the system's performance under high-cardinality label spaces. In Section \ref{experiment2}, we reconstructed the Dermnet dataset into Dermnet498, shifting from 23 broad super-classes to 498 granular sub-classes based on hierarchical metadata. Based on it, we systematically explored how SkinGPT-X maintains its stability as categorical complexity intensifies. We also curated a novel Rare Skin Disease Dataset (RSDD) based on the Rare Disease Diagnosis and Treatment Guidelines published by the National Health Commission of China in Section \ref{experiment3}. This benchmark served to stress-test the model's few-shot reasoning capabilities. Finally, to validate the core architectural innovation of our system, we conducted ablation studies and quantitative evaluations focusing on the EvoDerma-Mem in Section \ref{experiment4}. These experiments confirms the efficacy of the the proposed framework and the EvoDerma-Mem mechanism in synthesizing diagnostic rationales and evolving through historical case accumulation.

\subsection{SkinGPT-X Achieves State-of-the-Art Performance Against Four Baselines Across Four Datasets}
\begin{figure*}[ht]
    \centering
    \vspace{-10mm}  
    \includegraphics[width=0.97\linewidth]{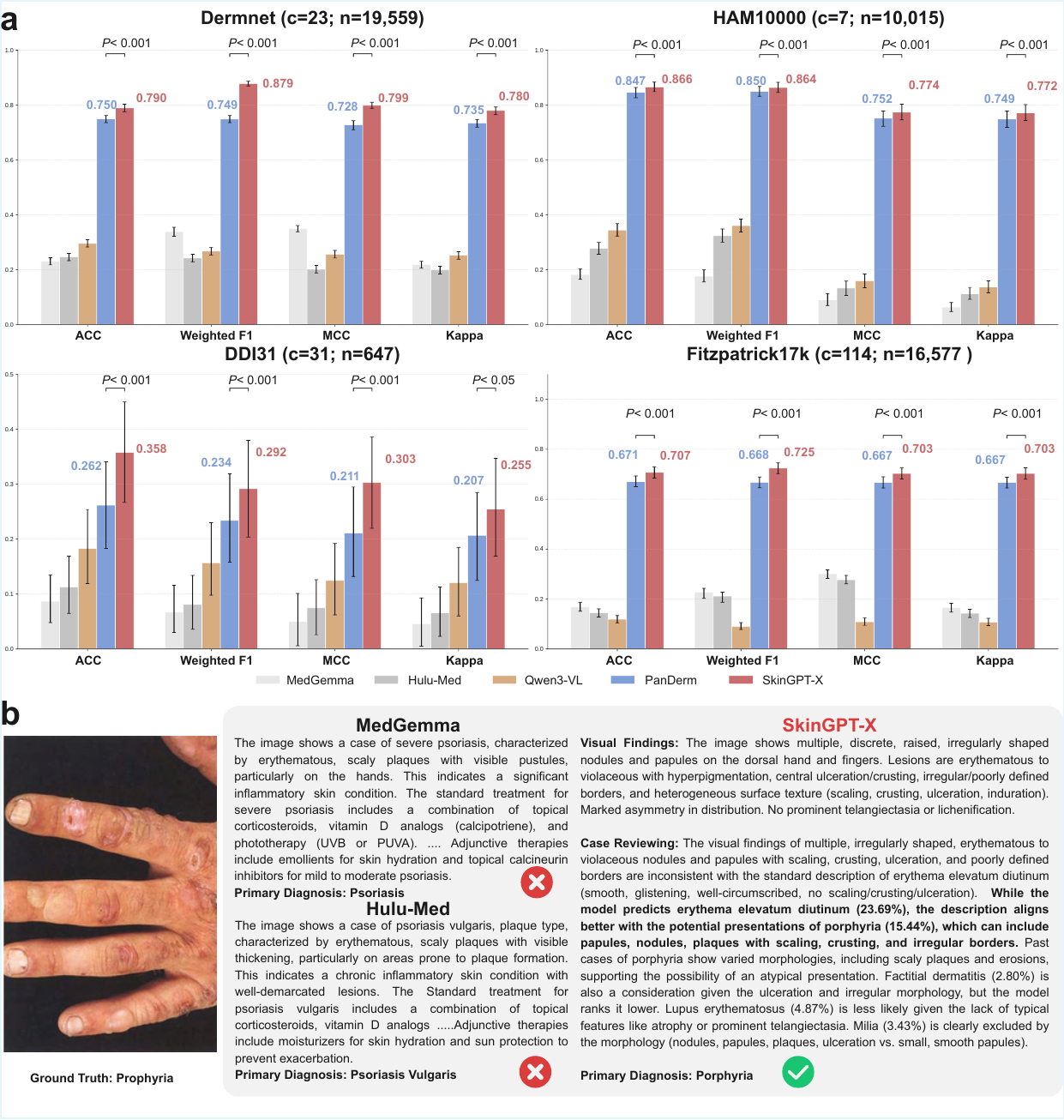} 
    \vspace{-3mm}
    \caption{\textbf{a,} Performance comparison of SkinGPT-X versus four state-of-the-art models (MedGemma, Hulu-Med, Qwen3-VL, and PanDerm) across four benchmark skin disease datasets. $c$ indicates the class number; $n$ represents the data size. Metrics include ACC, Weighted F1, MCC, and Cohen’s Kappa. The error bars represent 95\% CIs, and $P$ values were calculated using a two-sided $t$-test to indicate statistical significance between SkinGPT-X and the second-best model. \textbf{b,} Case study of diagnostic case reviewing and visual findings generation. The panels display the text-based outputs from MedGemma, Hulu-Med, and SkinGPT-X for representative clinical cases. The results illustrate that SkinGPT-X provides more reliable and transparent diagnostics by reviewing current clinical findings with its self-evolving agent memory.}
    \label{fig:experiments1}
\end{figure*}

\label{experiment1}
To evaluate the performance of SkinGPT-X, we conducted comprehensive comparative experiments across 4 benchmark datasets, including Dermnet, HAM10000, DDI31, and Fitzpatrick-17k. To provide a comprehensive view of their diagnostic efficacy, we assessed the models using 4 metrics: ACC, Weighted F1, MCC, and Cohen’s Kappa. Among these models, PanDerm was fine-tuned on each specific dataset for multi-class classification. In contrast, MedGemma\cite{sellergren2025medgemma}, Qwen3-VL\cite{qwen3technicalreport} and Hulu-Med\cite{jiang2025hulumedtransparentgeneralistmodel}, as general-purpose LLMs, conducted closed-ended dermatological assessments by selecting the most probable category from the label space of each dataset. As illustrated in Fig. \ref{fig:experiments1}a, SkinGPT-X consistently outperforms baseline models across all 4 evaluation metrics and datasets. In the Dermnet dataset ($n=19,559$), SkinGPT-X achieves a Weighted F1 of 87.9\% and an MCC of 79.9\%, significantly surpassing existing methods with high statistical confidence ($P < 0.001$). In the HAM10000 dataset ($n=10,015$), SkinGPT-X achieves the best performance in ACC (86.6\%), Weighted F1 (86.4\%), MCC (77.4\%), and Kappa (77.2\%). Notably, SkinGPT-X shows a substantial lead in the DDI31 dataset which typically presents a greater challenge due to diverse skin tones, reaching a Weighted F1 of 29.2\%, an ACC of 35.8\%, an MCC of 30.4\%, and a Kappa of 25.5\%, markedly higher than the leading baselines. Furthermore, in the Fitzpatrick-17k dataset ($n=16,577$), SkinGPT-X maintains its competitive edge, achieving the highest scores across all 4 metrics. These results collectively underscore SkinGPT-X's generalization and diagnostic accuracy across varying scales and complexities of dermatological data. 

\begin{figure*}[p]
    \centering
    \vspace{-12mm}  
    \includegraphics[width=0.94
    \linewidth]{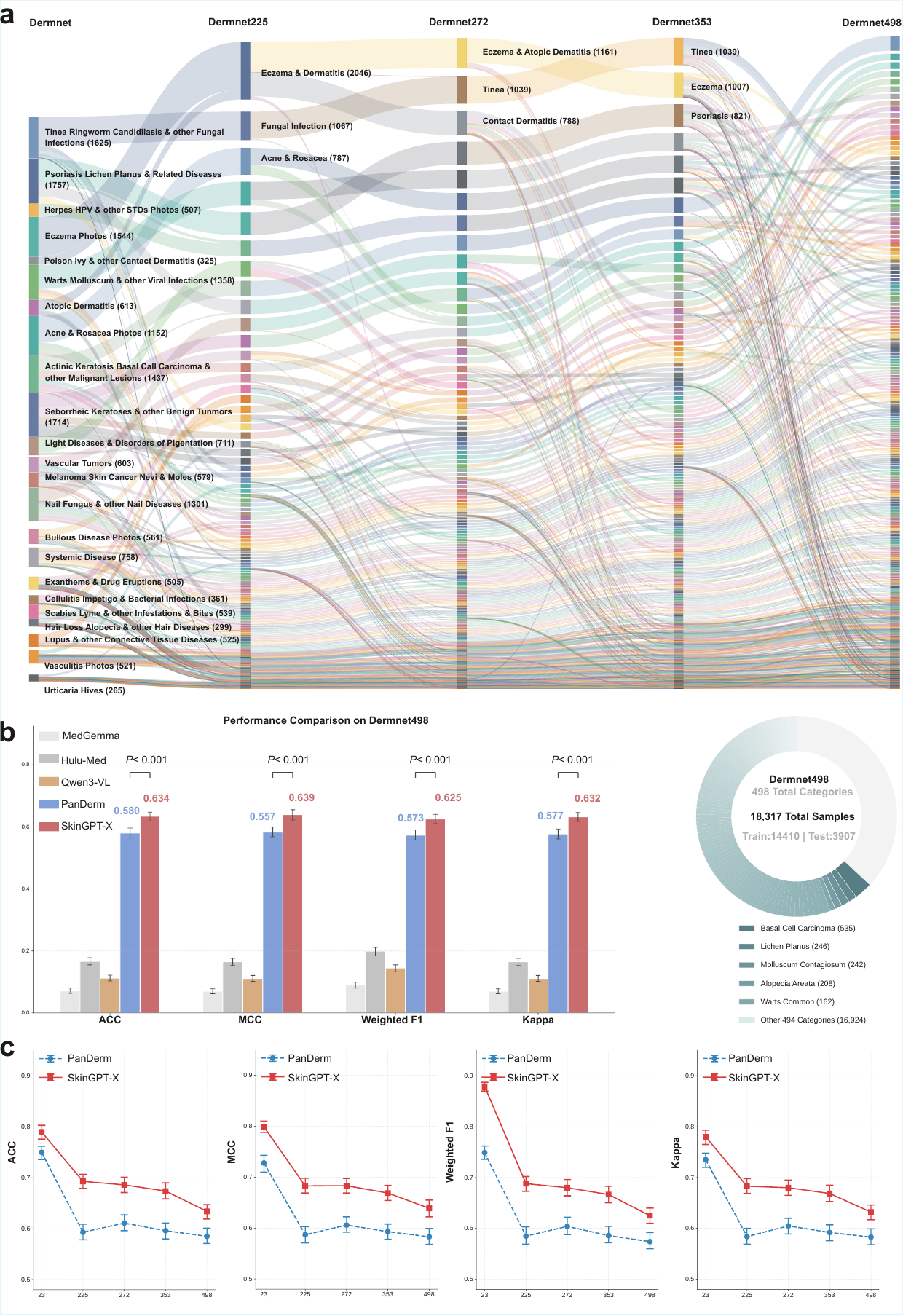} 
    \vspace{-5mm}  
    \caption{
    \textbf{a,} Data distribution and hierarchical composition of the  dataset series. This figure illustrates the incremental expansion of skin disease categories within the Dermnet dataset, scaling from Dermnet to the comprehensive Dermnet498. Each colored block represents a unique disease class. \textbf{b,} Comparative performance metrics on the Dermnet498 dataset. The bar chart presents ACC, MCC, Weighted F1, and Kappa scores for MedGemma, Hulu-Med, Qwen3-VL, PanDerm, and SkinGPT-X. Numerical values and error bars represent the mean and confidence intervals, with statistical significance between the top two models indicated by $P$-values. The radar chart displays the composition of Dermnet498. The legend specifies the top categories and their respective sample counts.\textbf{c,} Performance trends across varying category scales. The line graphs plot the changes in ACC, MCC, Weighted F1, and Kappa for the PanDerm and SkinGPT-X models as the number of disease categories increases.}
    \label{fig:experiments2}
\end{figure*}

\subsection{SkinGPT-X is Effective in High-Dimensional Classification Scenarios}
\label{experiment2}
As illustrated in Fig. \ref{fig:experiments2}a, while the original Dermnet dataset partitions samples into 23 categories, we found that these broad categories often grouped together clinically distinct diseases, leading to unclear definitions of specific conditions. Specifically, labels such as \textit{Psoriasis pictures Lichen Planus and related diseases} and \textit{Tinea Ringworm Candidiasis and other Fungal Infections} aggregate diverse conditions. To address, we introduced Dermnet498, a benchmark that re-categorizes the original images into 498 fine-grained classes based on their intrinsic, sample-specific metadata. This refinement process deconstructs broad categories into clinically  sub-classes. For instance, the category of \textit{Psoriasis pictures Lichen Planus and related diseases} is decoupled into 27 distinct classes, including 17 Psoriasis subtypes and 6 variants of Lichen Planus. Similarly, the Fungal Infection cluster is expanded into 24 specific categories, distinguishing site-specific infections such as CandidaAxillae and Candidasis Mouth. By shifting from 23 broad categories to 498 fine-grained categories, we establish a rigorous benchmark that mirrors the complexity of real-world dermatology.

As shown in Fig. \ref{fig:experiments2}b, this hierarchy encompasses a diverse distribution, ranging from common conditions to rare manifestations across 498 total categories. Detailed specifications of the Dermnet498 hierarchy are provided in Section \ref{Sec:Discussion_dataset}. 

As illustrated in Fig. \ref{fig:experiments2}c, we performed a comparative analysis against PanDerm by progressively increasing the diagnostic granularity from 23 to 498 categories. While the performance of both models naturally trends downward as the classification complexity intensifies, SkinGPT-X consistently maintains a robust margin over PanDerm. Notably, SkinGPT-X preserves its ACC, Weighted F1, and Cohen’s Kappa above the 60\% threshold, whereas the baseline performance drops more precipitously. 

As shown in the comparative benchmarks for Dermnet498 (Fig.~\ref{fig:experiments2}b),  MedGemma, Qwen3-VL, and Hulu-Med, demonstrate significant performance degradation when faced with such highly specialized and challenging fine-grained classification tasks. In contrast, the fine-tuned foundation model PanDerm achieves an ACC of 58.0\%, an MCC of 55.7\%, a Weighted F1 of 57.3\%, and a Kappa of 57.7\%. Remarkably, our SkinGPT-X system demonstrates a significant performance leap across all dimensions ($P < 0.001$). Specifically, it reaches an ACC of 63.4\% (+5.4\%), an MCC of 63.9\% (+8.2\%), a Weighted F1 of 62.5\% (+5.2\%), and a Kappa of 63.2\% (+5.5\%). This superior performance on Dermnet498 underscores SkinGPT-X's capacity for high-precision diagnosis. Its enhanced MCC and Weighted F1 suggest exceptional potential for the diagnosis of rare skin diseases, a domain where general LLMs lack sufficient specialized prior knowledge.
\begin{figure*}[t]
    \centering
    \vspace{-10mm}  
    \includegraphics[width=\linewidth]{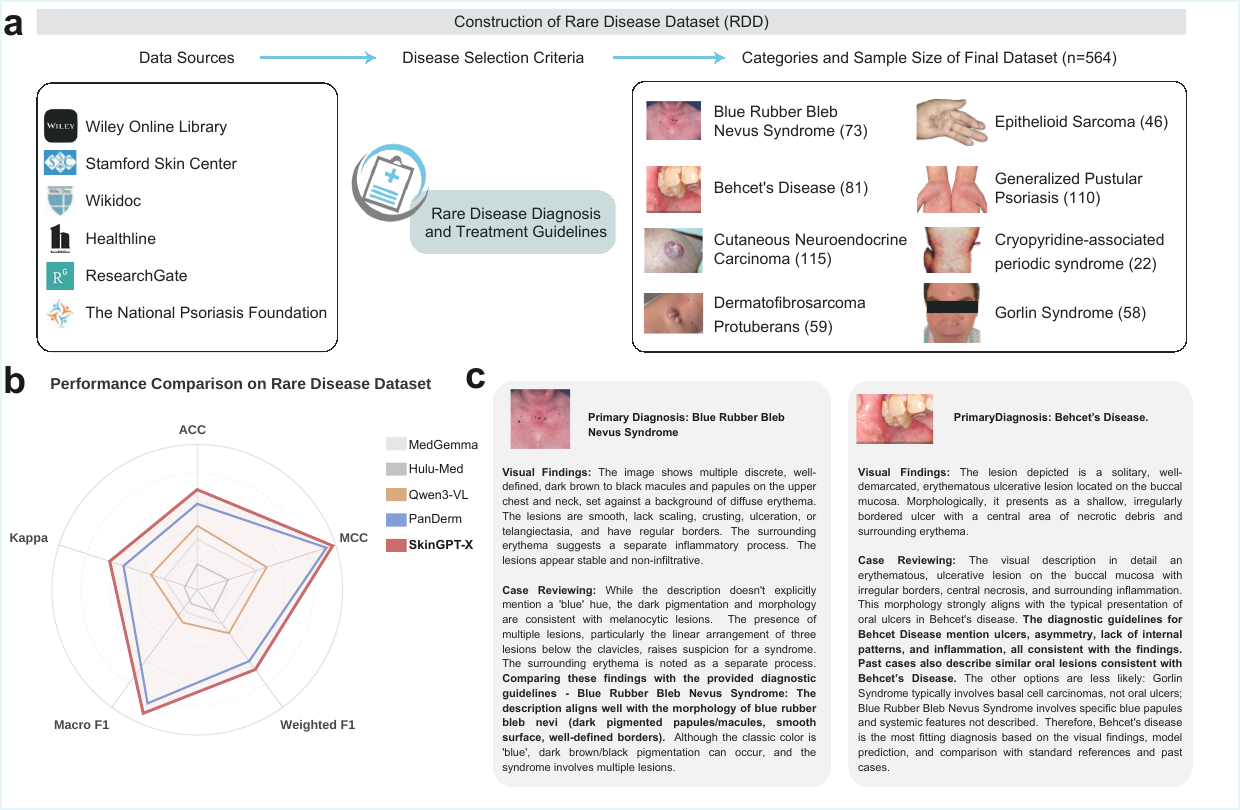} 
    \vspace{-4mm}  
    \caption{ 
    \textbf{a,} Construction pipeline of the Rare Skin Disease Dataset. The dataset integrates diverse data sources, including medical libraries, clinical centers, and specialized guidelines to ensure high-quality labels for rare dermatological conditions. The RSDD dataset comprises 8 rare conditions, including Cutaneous Neuroendocrine Carcinoma ($n=115$), Generalized Pustular Psoriasis ($n=110$), and Behcet's Disease ($n=81$), among others. 
    \textbf{b,} Multi-dimensional performance comparison on the RSDD ($n=564$). The radar chart illustrates the diagnostic efficacy of SkinGPT-X versus MedGemma, Hulu-Med, Qwen3-VL, and PanDerm across five metrics: ACC, MCC, Macro F1, Weighted F1, and Kappa.
    \textbf{c,} Representative case studies of rare disease diagnosis and differential reasoning. EvoDerma-Mem enhances the capability of the system differentiating the correct diagnosis form rare candidate skin diseases}
    \label{fig:experiments3}
\end{figure*}
\subsection{SkinGPT-X supports rare skin disease diagnosis}
\label{experiment3}
To further validate the robustness of SkinGPT-X in diagnosing rare and complex conditions where general LLMs often struggle due to the scarcity of training data, we constructed the Rare Skin Disease Dataset. As illustrated in Fig. ~\ref{fig:experiments3}a, the RSDD was rigorously curated by integrating diverse high-quality sources, including medical digital libraries, clinical centers, and specialized rare disease diagnosis and treatment guidelines. The final dataset comprises 564 samples across 8 distinct rare dermatological categories, with representative conditions such as Cutaneous Neuroendocrine Carcinoma ($n=115$), Generalized Pustular Psoriasis ($n=110$), and Behcet's Disease ($n=81$). 

As shown in the multi-dimensional performance comparison (Fig. ~\ref{fig:experiments3}b), SkinGPT-X achieves the highest diagnostic efficacy across 5 evaluated metrics, including ACC, MCC, Macro F1, Weighted F1, and Kappa. The radar chart visually demonstrates that SkinGPT-X not only significantly outperforms general LLMs like MedGemma and Hulu-Med which exhibit extremely limited prior knowledge in this domain, but also surpasses the fine-tuned foundation model PanDerm. This superiority underscores the effectiveness of our proposed SkinGPT-X system in harmonizing visual features with specialized clinical knowledge. 


\begin{figure*}[htp]
    \centering
    \vspace{-10mm}  
    \includegraphics[width=\linewidth]{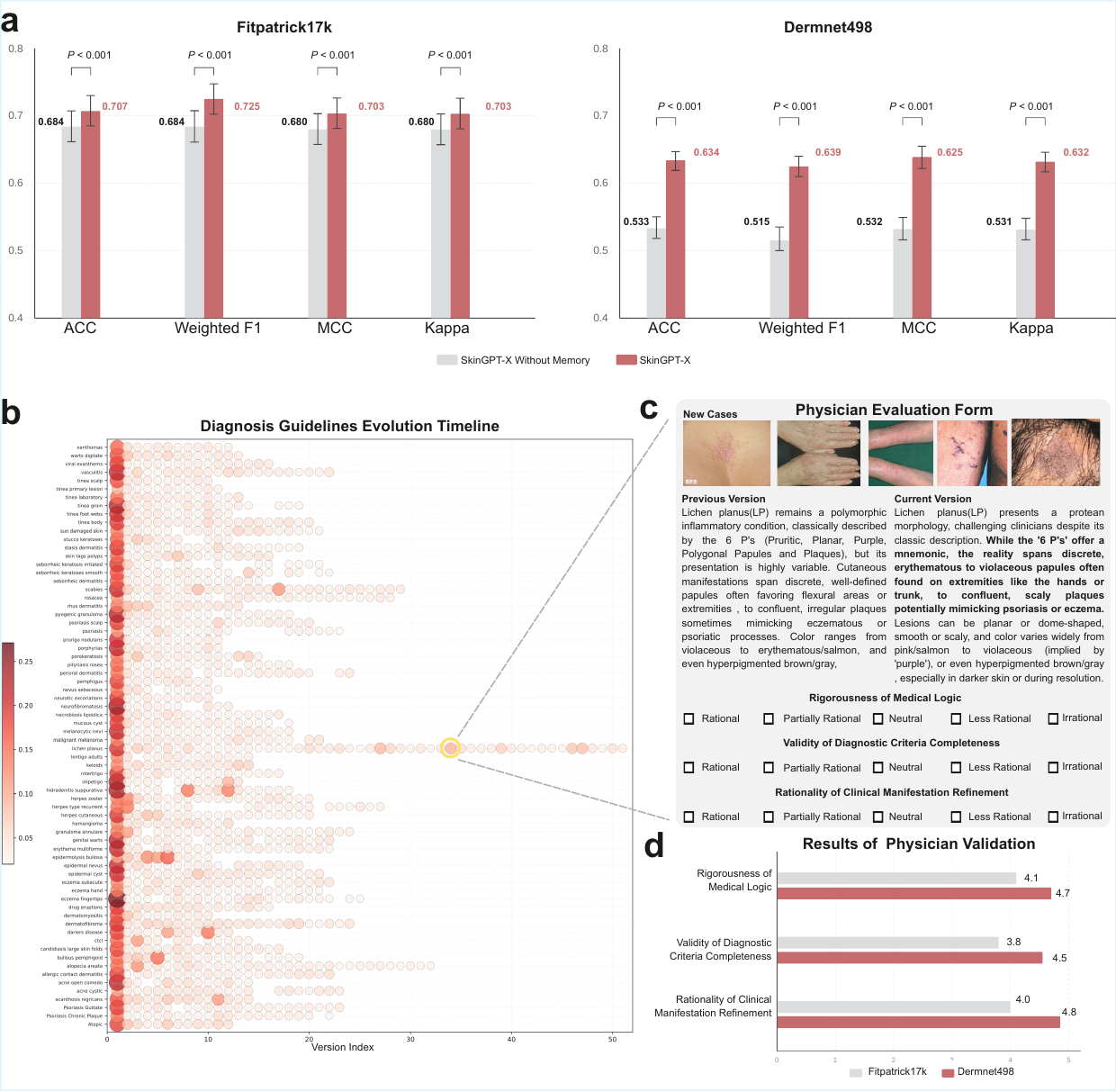} 
    \caption{
    \textbf{a,} Performance comparison between SkinGPT-X and the baseline framework (without memory) on Fitzpatrick-17k and Dermnet498 datasets. Data are presented as mean metrics with 95\% confidence intervals; statistical significance was determined via two-sided t-tests ($p < 0.001$).
    \textbf{b,} Spatiotemporal evolution trajectory of diagnostic guidelines. The bubble color intensity represent degree of knowledge refinement, where deeper shades indicate more substantial updates to the diagnostic guidelines.
    \textbf{c,} The standardized Physician Evaluation Form used for blinded clinical review, focusing on Rigorousness of Medical Logic, Validity of Diagnostic guidelines Completeness and Rationality of Clinical Manifestation Refinement.
    \textbf{d,} Results of physician validation across three critical dimensions.}
    \label{fig:experiments4}
\end{figure*}
\subsection{EvoDerma-Mem is Indispensable for Evolving Clinical Reasoning and Expert-Level Knowledge Maturation}
\label{experiment4}
EvoDerma-Mem enhances the diagnostic process by providing evolved diagnostic guidelines and retrieved past cases. As illustrated in Fig. \ref{fig:experiments3}c, for the case of Blue Rubber Bleb Nevus Syndrome,  by cross-referencing the evolved guidelines, the system recognized that \textit{while the classic color is blue, dark pigmentation is a known variant, especially when presented as multiple, smooth-surfaced lesions in a linear arrangement}. This allowed the model to correctly identify a systemic syndrome rather than dismissing the findings as isolated melanocytic lesions. Similarly, in the diagnosis of Behcet’s Disease, the system analyzed a solitary, erythematous ulcerative lesion on the buccal mucosa. While such a finding could be non-specific, the retrieved past cases and guidelines from EvoDerma-Mem emphasized the importance of \textit{shallow, irregular borders with central necrotic debris and surrounding erythema.} By comparing these visual findings against historical cases, the system was able to morphologically similar disorders such as Gorlin Syndrome an Blue Rubber Bleb Nevus Syndrome.

EvoDerma-Mem emulates the iterative learning trajectory of a senior clinician. This continuous refinement is visualized in the Guidelines Evolution Timeline (Fig.~\ref{fig:experiments4}b), where each node represents a specific iteration of a disease's diagnostic guidelines. The size and color intensity of each bubble quantify the magnitude of descriptive refinement, relative to its previous version. A salient example is the evolution of Lichen Planus (LP) guidelines as shown in Fig.~\ref{fig:experiments4}c. While the previous version was confined to the classic \textit{6 P’s} mnemonic (Pruritic, Planar, Purple, Polygonal, Papules, and Plaques), the current version has matured into a high-fidelity clinical manifest. By assimilating new cases, the system has extended its descriptive scope to include atypical anatomical sites (e.g., trunk and hands) and identified critical morphological mimics, such as \textit{dome-shaped} lesions and \textit{scaly plaques} that frequently overlap with psoriasis or eczema.

To evaluate the critical necessity of EvoDerma-Mem, we conducted ablation studies and clinical expert validations. As illustrated in Fig.~\ref{fig:experiments4}a, ablation studies demonstrate that the integration of EvoDerma-Mem consistently yields performance gains across five metrics (ACC, BACC, Weighted F1, MCC, and Kappa, all $p < 0.001$).Specifically, on the Fitzpatrick17k dataset, the inclusion of evolved guidelines and historical case retrieval elevated the ACC from 68.4\% to 70.7\%, and the Weighted F1 from 68.4\% to 72.5\%, representing a significant improvement in diagnostic precision. Similar trends were observed across other metrics, underscoring the system's enhanced reliability in dermatological diagnosis.Results on the Dermnet498 dataset further validate these findings, where the performance gains were even more pronounced. The full SkinGPT-X framework achieved superior performance compared to the version without memory, with an +10.1\% improvement in ACC, a +11.0\% improvement in Weighted F1, a +10.7\% improvement in MCC, and a +10.1\% improvement in Kappa. These results indicate that EvoDerma-Mem significantly enhances the framework's ability to handle diverse pathological presentations, particularly for high-dimensional categories with limited training samples.

To assess the practical clinical utility of the evolved knowledge, we invited two senior dermatologists to conduct a blinded clinical review of 600 evolutionary cases. As illustrated in Fig. \ref{fig:experiments4}c The experts evaluated the EvoDerma-Mem system from three critical dimensions: 1) Rigorousness of Medical Logic; 2) Validity of Diagnostic guidelines Completeness; 3) Rationality of Clinical Manifestation Refinement.

EvoDerma-Mem achieves consistently excellent performance across both the Fitzpatrick-17k and Dermnet498 datasets. In the quantitative evaluations, the system maintained exceptional scores on a 5-point scale across all three critical dimensions, achieving 4.3 and 4.7 for Medical Logic Rigorousness, 3.8 and 4.5 for Diagnostic guidelines Completeness, and 4.0 and 4.8 for Clinical Manifestation Refinement on the Fitzpatrick-17k and Dermnet498 benchmarks, respectively. These results underscore that EvoDerma-Mem is not merely a supplementary module but the indispensable architectural foundation that enables SkinGPT-X to maintain professional rigor and descriptive depth in real-world clinical workflows.

\section{Discussion}

By emulating the cognitive pathways of clinical experts, SkinGPT-X effectively achieves a cognitive segregation of visual findings extraction from diagnostic decision-making. By retrieving standardized medical knowledge from the Skin Handbook at this critical junction, the system provides a rigorous clinical anchor to validate the Disease Diagnosis Agent's hypothesis for candidate diseases \cite{zakka2024aloha}. Beyond static data, the EvoDerma-Mem mechanism allows the system to accumulate actual clinical experience, constantly refining its diagnostic guidelines through various historical cases. The system conducts a multi-dimensional alignment between current visual findings and historical cases. By synergizing book-based dermatological knowledge with dynamically evolved diagnostic guidelines, SkinGPT-X performs a rigorous validation of the initial hypothesis to identify the most clinically consistent diagnosis. This workflow enhances both the traceability and interpretability of the reasoning process. Through multi-agent synergistic verification and cross-logical associations, the system attains superior diagnostic reliability that avoids the 'black box' limitations of monolithic large-scale models \cite{moor2023foundation}. Simultaneously, the EvoDerma-Mem mechanism emulates the cognitive evolution of senior dermatologists, which allows the system to supplement foundational medical knowledge with a more comprehensive layer of dermatological insights, ultimately facilitating highly granular diagnostics and significantly improving the efficiency of identifying atypical or rare clinical presentations.

Memory has emerged as a pivotal frontier in the evolution of AI agents \cite{wang2024survey_agents}. Within the domain of dermatology, most diagnostic systems currently rely on static RAG to provide agents with foundational medical knowledge. However, generalized medical knowledge often proves insufficient for dealing with complex clinical cases \cite{jabbour2023deep}. Furthermore, fine-tuning LLMs is frequently constrained by the limitations of few-shot learning; with sparse clinical data, these models often fail to capture specific, discriminative diagnostic features \cite{vinyals2016matching}. To address this, we designed the EvoDerma-Mem mechanism, which emulates the empirical synthesis process of experienced clinicians by continuously distilling standout details from novel cases to form evolving diagnostic guidelines. During the diagnostic phase, EvoDerma-Mem provides agents with both visually similar historical cases and refined diagnostic guidelines. This dynamically evolving agent memory offers superior heuristic guidance even in few-shot scenarios, while simultaneously ensuring a more transparent and traceable diagnostic rationale by explicitly linking current findings to retrieved historical cases and their synthesized diagnostic guidelines \cite{lewis2020retrieval}.

The experimental results further illuminate the effectiveness of the EvoDerma-Mem mechanism, particularly in stabilizing the reasoning process as diagnostic complexity scales. As illustrated in Fig. \ref{fig:experiments2}c, while baseline models suffered significant performance degradation when transitioning to fine-grained classification (up to 498 classes), SkinGPT-X maintained high diagnostic resilience, outperforming specialized foundation models like PanDerm \cite{zhou2023skingpt4}. This advantage is even more pronounced in the RSDD, where the system’s ability to engage in guidelines-aligned inference allows it to transcend simple pattern matching, effectively addressing the challenges of few-shot clinical reasoning. As evidenced in Fig.
\ref{fig:experiments4}, both the ablation studies and the double-blind reviews conducted by senior dermatologists substantiate that EvoDerma-Mem is an indispensable architectural component. The ablation of this mechanism results in a statistically significant degradation in diagnostic precision. The evolved guidelines generated by the system exhibit a high degree of feature comprehensiveness and inferential coherence, demonstrating a rigorous alignment with established professional clinical standards \cite{kelly2019key_eval}.

Despite the significant advancements demonstrated by SkinGPT-X, several intrinsic limitations regarding its architectural complexity and clinical deployment warrant further discussion. First, the transition from monolithic models to a MAS framework introduces a necessary trade-off between reasoning depth and computational efficiency. The collaborative orchestration required for visual feature extraction, EvoDerma-Mem retrieval, and iterative guideline synthesis results in prolonged processing times, which may pose challenges in high-throughput clinical settings where real-time feedback is critical \cite{kasneci2023chatgpt_healthcare}. Furthermore, the system’s diagnostic reliability is sensitive to the heterogeneity of image acquisition across different medical centers. Variations in sensor calibration, ambient lighting, and magnification levels can induce a distribution shift in the latent feature space, potentially leading to device-specific biases that degrade the retrieval accuracy of the EvoDerma-Mem mechanism \cite{guolam2021skin_bias}. In future research, we will focus on optimizing agent communication protocols and implementing robust domain-adaptation techniques to ensure both operational speed and cross-center generalizability.

This study establishes a new paradigm for dermatological artificial intelligence by combining the self-evolving agent memory with systematic collaboration of multiple AI agents. The incorporation of the EvoDerma-Mem mechanism and guidelines-aligned reasoning strategies enhances diagnostic resilience, particularly in few-shot scenarios involving rare skin diseases, while simultaneously ensuring a more transparent and traceable diagnostic rationale. These findings suggest that the future of medical AI lies not merely in the expansion of static knowledge bases, but in the explicit emulation of clinical experience and the structural stabilization of deep-reasoning workflows.
\section{Methods}
\subsection{Metrics}
\textcolor{black}{To comprehensively evaluate the performance of the proposed model, we employ five widely recognized metrics: ACC, Macro F1, Weighted F1, MCC, and Cohen's Kappa. These metrics are derived from the confusion matrix components: True Positives ($TP$), True Negatives ($TN$), False Positives ($FP$), and False Negatives ($FN$).}

\noindent \textcolor{black}{\textbf{ACC} measures the proportion of correctly classified instances among the total number of cases, providing a general assessment of the model's performance:
\begin{equation}
    ACC = \frac{TP + TN}{TP + TN + FP + FN}
\end{equation}}

\noindent \textcolor{black}{\textbf{Macro F1} is the arithmetic mean of class-specific F1 values. It assigns equal weight to each class, ensuring that performance on minority classes is not overshadowed by the majority class in imbalanced datasets:
\begin{equation}
Macro\ F1 = \frac{1}{2} \left( \frac{2TP}{2TP + FP + FN} + \frac{2TN}{2TN + FN + FP} \right)
\end{equation}}

\noindent \textcolor{black}{\textbf{Weighted F1} averages the F1-score of each class, weighted by their respective number of samples ($N_i$):
\begin{equation}
    Weighted\ F1 = \sum_{i=1}^{L} \left( \frac{2 \cdot Precision_i \cdot Recall_i}{Precision_i + Recall_i} \cdot \frac{N_i}{N_{total}} \right)
\end{equation}}

\noindent \textcolor{black}{\textbf{MCC} is a reliable statistical rate that produces a high score only if the prediction performs well across all categories. To fit the column width, the formula is expressed as:
\begin{equation}
\begin{split}
    MCC = & (TP \cdot TN - FP \cdot FN) / \\
          & \sqrt{(TP+FP)(TP+FN)(TN+FP)(TN+FN)}
\end{split}
\end{equation}}

\noindent \textcolor{black}{\textbf{Cohen's Kappa} measures the agreement between predicted and observed classifications while accounting for the possibility of agreement by chance:
\begin{equation}
    \kappa = \frac{p_o - p_e}{1 - p_e}
\end{equation}
where $p_o$ is the observed agreement and $p_e$ is the expected agreement due to chance.}

\subsection{Dataset}\label{Sec:Discussion_dataset}
To evaluate the diagnostic efficacy and generalization of SkinGPT-X, we employed five benchmark datasets and curated a specialized rare disease evaluation suite:

\textbf{Dermnet}, one of the largest and most widely recognized public clinical image benchmarks for skin disease classification. This dataset comprises 18,856 images specifically curated to represent the diversity of dermatological conditions encountered in clinical practice. The images are organized into 23 primary super-classes, covering a broad spectrum of common and complex pathologies.

\textbf{HAM10000}, which stands as one of the most prominent and frequently cited public dermoscopic image benchmarks for the classification of pigmented skin lesions. This dataset comprises 10,015 high-quality dermoscopic images, meticulously curated to address the challenges of automated diagnosis in dermatology. The images are organized into 7 primary diagnostic categories, representing a significant spectrum of common benign and malignant pigmented pathologies.

\textbf{Fitzpatrick-17k}, a large-scale clinical image benchmark specifically developed to address the critical need for skin tone diversity and algorithmic fairness in dermatological AI. This dataset comprises 16,577 clinical images curated to represent a wide variety of skin conditions across all six Fitzpatrick skin types (FST I to VI). Each image is meticulously annotated with both a skin condition label and its corresponding skin type, making it a pivotal resource for studying demographic bias.

\textbf{DDI31}, a pioneering benchmark specifically designed to evaluate the fairness and performance of dermatology AI models across diverse skin tones. The dataset contains a total of 656 images, meticulously curated and pathologically confirmed from Stanford Clinics. In our study, we partitioned the dataset into a training set and a testing set using a $4:1$ ratio to ensure a balanced evaluation. To accommodate a wide range of dermatological conditions, the dataset is categorized into 31 distinct diagnostic classes, encompassing both benign and malignant skin lesions.

\textbf{Dermnet225} is the most aggressively consolidated variant among the sub-datasets, featuring a highly curated set of 225 classes. The mapping logic for Dermnet225 prioritizes clinical grouping to minimize noise from ambiguous sub-labels. We implemented a multi-tiered infection classification, where various fungal species were grouped under \textit{Fungal Infection} and diverse bacterial presentations (e.g., \textit{Folliculitis}, \textit{Impetigo}, and \textit{Cellulitis}) were unified into \textit{Bacterial Infection}. Similarly, benign soft tissue growths and different types of viral warts were merged based on their biological commonalities. This dataset serves as a benchmark for evaluating models on a consolidated yet diverse label space.

\textbf{Dermnet272} represents a more condensed hierarchical reorganization, targeting a classification space of 272 categories. The construction of this dataset focused on reducing fragmented sub-classifications by applying deeper medical grouping logic. Beyond the standard class consolidation used in Dermnet353, we implemented an advanced merging logic to handle varied spellings of complex conditions (e.g., merging multiple variants of \textit{Ichthyosis}). Furthermore, various insect bites and stings were grouped into a unified \textit{Bites and Stings} class, and miscellaneous nail disorders were aggregated to ensure each category maintained sufficient diagnostic representativeness.

\textbf{Dermnet353} is a refined intermediate-scale version of the dataset, designed to balance taxonomic breadth with categorical distinctness. While the original Dermnet is structured into 23 super-classes, we implemented a rule-based merging strategy to map the underlying fine-grained images into 353 clinically relevant categories. This process involved normalizing labels and consolidating sub-categories that share high morphological similarity, such as merging various types of Contact Dermatitis (allergic, irritant, and phytophotodermatitis) into a single group. Additionally, parasitic infestations like Scabies and Pediculosis were unified due to their overlapping clinical presentation in a computer vision context.

\textbf{Dermnet498}, is a high-cardinality benchmark we constructed by refining the original Dermnet dataset. While the original Dermnet is organized into 23 broad super-classes, it contains rich, hierarchical metadata where images are associated with specific clinical sub-categories. Leveraging this hierarchy, we reorganized the dataset by re-mapping the images from 23 coarse super-classes to 498 distinct fine-grained sub-classes. During the curation process, we performed a rigorous data cleaning step: images associated with sub-labels that lacked clinical significance or were diagnostically ambiguous were excluded. Consequently, the final \textbf{Dermnet498, Dermnet353, Dermnet272, Dermnet225} datasets comprise 18,853 images, a slight reduction from the original count. This transformation shifts the objective from broad-category classification to a more challenging fine-grained recognition task within an extensive label space. 

The \textbf{Dermnet498} dataset was partitioned into a training set of 14,856 images and a test set of 3,997 images, ensuring that both sets encompass the full spectrum of the 498 sub-classes.

\subsection{The EvoDerma-Mem Mechanism in SkinGPT-X}
To simulate the clinical expertise accumulation process of a dermatologist, who refines diagnostic experience through continuous clinical practice. We developed a novel self-evolving agent memory name EvoDerma-Mem for dermatological diagnosis, as illustrated in Fig. \ref{fig:skingpt_framework}(b). This mechanism comprises a structured database and a knowledge evolution process.
\subsubsection{Case Representation and agent memory Storage}
Initially, each historical sample $S_i$ in the repository is processed through a pre-trained Feature Extractor $\Phi$. This module encodes the high-dimensional semantic features of the clinical image $I_i$ into a compact embedding code $\bm{z}_i$:
\begin{equation}
    \bm{z}_i = \Phi(I_i), \quad \bm{z}_i \in \mathbb{R}^d
\end{equation}
Each memory entry $M_i$ is stored as a linked triplet within a graph database, formally defined as:
\begin{equation}
    M_i = \langle \bm{z}_i, K_i, D_i \rangle
\end{equation}
where $K_i$ represents the textual \textit{Key Findings} and $D_i$ denotes the corresponding expert diagnosis. During inference, the system retrieves the most relevant historical cases by calculating the cosine similarity between the query embedding $\bm{z}_{q}$ and the stored embeddings:
\begin{equation}
    \text{Score}(M_q, M_i) = \frac{\bm{z}_q \cdot \bm{z}_i}{\|\bm{z}_q\| \|\bm{z}_i\|}
\end{equation}
\subsubsection{Knowledge Synthesis and Guideline Evolution}
The knowledge evolution process leverages a specialized \textbf{Summarization Agent} $\mathcal{A}$ to synthesize collective diagnosis guidelines. For a specific disease category $C$, the agent analyzes the set of all associated findings $\mathcal{K}_C = \{K_1, K_2, \dots, K_n\}$ to generate the initial diagnostic guidelines $G_C^0$:
\begin{equation}
    G_C^0 = \mathcal{A}(\mathcal{K}_C)
\end{equation}

The system implements a dynamic evolution process triggered by an accumulation threshold $N_{thresh}$. When the number of new samples $\Delta N \geq N_{thresh}$, the Summarization Agent updates the current guidelines $G_C^t$ to the next iteration $G_C^{t+1}$ by integrating the latest findings $\mathcal{K}_{new}$:
\begin{equation}
    G_C^{t+1} = \mathcal{A}(G_C^t \oplus \mathcal{K}_{new})
\end{equation}
where $\oplus$ represents the knowledge integration operation that resolves contradictions and incorporates novel morphological evidence. This closed-loop evolution ensures that SkinGPT-X continuously refines its clinical reasoning logic, achieving a progressive maturation of diagnostic intelligence as the longitudinal dataset grows.


\subsection{Multi-Agent Collaborative Diagnosis Framework}

To achieve the transparent and trustworthy dermatological reasoning, SkinGPT-X implements a collaborative framework comprising multiple specialized agents as shown in Algorithm \ref{alg:skingpt_inference}. 
Algorithm \ref{alg:agent_details_math} provides the comprehensive computational procedures for both the autonomous agents and the knowledge-retrieval functions within the framework.
\begin{algorithm}[H]
\caption{SkinGPT-X Inference Process}
\label{alg:skingpt_inference}
\begin{algorithmic}[1]
\Require Input patient image $\mathcal{I}$, Skin handbook $\mathcal{B}$, Agent memory $\mathcal{M}$ (containing historical cases $\mathcal{K}_{hist}$ and evolved guidelines $\mathcal{G}$).
\Ensure Final validated diagnostic report $\mathcal{R}^*$ and optimal diagnosis $D^*$.

\State $\bm{z}_{\mathcal{I}} \leftarrow \text{VisionEncoder}(\mathcal{I})$ \hfill \Comment{Generate visual embedding}
\State $\mathcal{P}_{vis} \leftarrow \text{VisualAgent}(\text{Prompt}_{morph}, \mathcal{I})$ \hfill \Comment{Objective morphological}
\State $\mathcal{D}_{pre} \leftarrow \text{Pre-DiagAgent}(\mathcal{I})$ \hfill \Comment{Top-5 candidate diseases}

\For{each $d_i \in \mathcal{D}_{pre}$}
    \State $\mathcal{K}_{prior, i} \leftarrow \text{Retrieve}(\mathcal{B}, d_i)$ \hfill \Comment{Fetch textbook standards}
    \State $\mathcal{G}_{evo} \leftarrow \bigcup_{d_i \in \mathcal{D}_{pre}} \mathcal{G}[d_i]$ \hfill \Comment{Self-evolved guidelines}
\EndFor
\State $\mathcal{K}_{hist} \leftarrow \text{Query}(\mathcal{M}, \bm{z}_{\mathcal{I}})$ \hfill \Comment{Retrieve top-5 similar cases}
\State $\mathcal{X}_{rev} \leftarrow \{ \mathcal{P}_{vis}, \mathcal{D}_{pre}, \mathcal{K}_{prior}\}$ \hfill \Comment{Construct evidence space}
\State $D^* \leftarrow \mathcal{A}_{rev}(\mathcal{X}_{rev}, \mathcal{K}_{hist}, \mathcal{G}_{evo} )$ \hfill \Comment{Cross-verification, details are listed in Section \ref{sec:thinkingprocess}}

\State $\mathcal{R}^* \leftarrow (D^*, \mathcal{P}_{vis})$ 
\State \Return \text{Validated Report} $\mathcal{R}^*$
\end{algorithmic}
\end{algorithm}

\subsubsection{Independent Visual and Diagnostic Agents}
The reasoning pipeline begins with the \textbf{Vision Agent}, powered by the Qwen3-VL model. Rather than providing a definitive diagnosis, this agent is restricted to the objective characterization of pathological features from the input imagery. It generates a structured description of the lesion's morphology, pigmentation patterns, and boundary conditions, establishing a factual foundation for downstream reasoning.

Subsequently, the \textbf{Disease Diagnosis Agent} utilizes a scalable architecture designed for modular integration. This module can embed one or more dermatology-specific foundation models as pluggable components. The agent synthesizes the visual evidence to output the top-5 candidate diagnoses $\mathcal{D} = \{d_1, \dots, d_5\}$, where each diagnosis $d_i$ is associated with a corresponding confidence score $P(d_i)$ for $i \in \{1, \dots, 5\}$.

\subsubsection{Knowledge-Augmented Reasoning via RAG}
To reinforce diagnostic rigor, we integrate a RAG module supported by a custom-built Skin Handbook. This knowledge base $\mathcal{B}$ was constructed by extracting domain-specific expertise from the Oxford Handbook of Medical Dermatology\cite{griffiths2020oxford}, covering the structure and function of the skin, standardized diagnostic standards, and distinguish features of diverse skin conditions. 

The RAG module retrieves specific medical standards for each candidate disease:
\begin{equation}
    \mathcal{K}_{prior} = \text{Retrieve}(\mathcal{B}, d_i)
\end{equation}

\begin{algorithm}[H]
\caption{Details of the computational procedures for agents and the knowledge-retrieval functions}
\label{alg:agent_details_math}
\begin{algorithmic}[1]
\Require 
    Input patient image $\mathcal{I}$; 
    Skin handbook $\mathcal{B}$; 
    Agent memory $\mathcal{M}$ (containing historical cases $\mathcal{K}_{hist}$ and evolved guidelines $\mathcal{G}_{evo}$);

\Statex
\Statex \textbf{Function} $\text{Embedding}(\mathcal{I})$:
\State \quad $\bm{z}_{\mathcal{I}} = \text{PanDerm}_{\text{frozen}}(\mathcal{I})$ 
\State \quad \Return $\bm{z}_{\mathcal{I}} \in \mathbb{R}^d$. \hfill \Comment{Visual Embeddings of $\mathcal{I}$}

\Statex
\Statex \textbf{Function} $\text{VisualAgent}(\text{Prompt}_{morph}, \mathcal{I})$:
\State \quad $\mathcal{P}_{vis} = \text{LLM}_{\text{Qwen3-VL}}(\text{Prompt}_{morph} \oplus \mathcal{I})$ 
\State \quad \Return $\mathcal{P}_{vis}$; \hfill \Comment{Morphological descriptions of $\mathcal{I}$}

\Statex
\Statex \textbf{Function} $\text{Pre-DiagnosisAgent}(\mathcal{I})$:
\State \quad $\mathcal{P}(d \mid \mathcal{I}) = \text{Softmax}(\text{PanDerm}_{\text{tuned}}(\mathcal{I}))$ 
\State \quad $\mathcal{D}_{pre} = \{ (d_i, p_i) \mid \text{rank}(P(d_i \mid \mathcal{I})) \le 5 \}$
\State \quad \Return $\mathcal{D}_{pre}$ \hfill \Comment{Diagnostic candidate set, where $d_i$ is the disease label and $p_i$ is the confidence score;}

\Statex
\Statex \textbf{Function} $\text{Retrieve}(\mathcal{B}, d_i)$:
\State \quad $\mathcal{K}_{prior} = \arg\max_{k \in \mathcal{B}} \cos( \text{Emb}_{text}(d_i), \text{Emb}_{text}(k) )$ 
\State \quad \Return $\mathcal{K}_{prior}$ \hfill \Comment{Medical standards from textbook}

\Statex
\Statex \textbf{Function} $\text{Query}(\mathcal{M}, \bm{z}_{\mathcal{I}})$:
\State \quad $\mathcal{J}^* = \text{TopK}_{j} \left( \frac{\bm{z}_{\mathcal{I}} \cdot \bm{z}_j}{\|\bm{z}_{\mathcal{I}}\| \|\bm{z}_j\|} \right), \text{ s.t. } (\bm{z}_j, d_j) \in \mathcal{M}, K=5$ 
\State \quad $\mathcal{K}_{hist} = \{ (d_j, \mathcal{P}_j) \}_{j \in \mathcal{J}^*}$
\State \quad \Return $\mathcal{K}_{hist}$  \hfill \Comment{Similar historical cases from agent momery}

\end{algorithmic}
\end{algorithm}

\subsubsection{The Enhanced Case Reviewing based on Self-Evolving Agent Memory}
\label{sec:thinkingprocess}
To achieve high-fidelity diagnostic conclusions, we design a \textbf{Case Reviewing} agent $\mathcal{A}_{rev}$ powered by the Qwen3-A30\cite{qwen3technicalreport}. This module functions as a clinical reviewer, performing a traceable review of multi-source evidence to emulate the rigorous synthesis process of senior dermatologists.

First, diverse clinical information mentioned above is collected and organized to provide a unified basis for multi-dimensional cross-verification. Formally, for a given sample $q$, the input space $\mathcal{X}_{rev}$ is defined as:
\begin{equation}
    \mathcal{X}_{rev} = \{ \mathcal{P}_{vis}, \mathcal{D}_{pre}, \mathcal{K}_{prior}\}
\end{equation}
where:
\begin{itemize}
    \item $\mathcal{P}_{vis}$ represents the \textbf{Multimodal Perception}, including visual pathological descriptions and key findings.
    \item $\mathcal{D}_{pre} = \{ (d_i, p_i) \}_{i=1}^5$ is the list of \textbf{Candidate Diagnoses} with their respective confidence scores.
    \item $\mathcal{K}_{prior} = \text{Retrieve}(\mathcal{B}, d_i)$ denotes the \textbf{Expert Knowledge Alignment}, retrieving standardized standards from the Skin Handbook $\mathcal{B}$.
\end{itemize}

Second, the agent evaluates the inferential coherence across multiple information dimensions to derive the optimal diagnosis $D^*$. Unlike conventional models that rely on frozen parameters $\theta$, our system leverages the in-context learning capabilities of the LLM to orchestrate foundational medical knowledge with experiential clinical insights:

\begin{equation}
    D^* \leftarrow \mathcal{A}_{\text{val}}(\mathcal{X}_{rev}, \mathcal{K}_{\text{hist}}, \mathcal{G}_{\text{evo}})
\end{equation}
where $\mathcal{K}_{hist}$ and $\mathcal{G}_{\text{evo}}$ are retrieved from the agent memory. $\mathcal{K}_{hist}$ represents the \textbf{Historical Cases}, containing the top-5 similar cases' visual key findings and their diagnosis. $\mathcal{G}_{\text{evo}}$ signifies the self-evolved diagnostic guidelines refined by the Summarize Agent. By incorporating $\mathcal{G}_{\text{evo}}$, the system transcends simple instance-based retrieval, leveraging high-level clinical abstractions synthesized from collective historical encounters. This mechanism ensures that the final output is not a mere statistical prediction, but a clinically validated conclusion grounded in the synergistic verification of textbook standards and empirical precedents.

The reasoning process is operationalized through a five-stage \textit{Review-and-Synthesis} protocol:

\begin{itemize}
    \item \textbf{Stage 1: Visual Feature Validation.} The agent first performs an independent review of the initial perception $\mathcal{P}_{\text{vis}}$. It identifies and rectifies potential omissions in morphological descriptors to ensure a high-fidelity visual foundation.
    \item \textbf{Stage 2: Canonical Guidelines Cross-Check.} The validated findings are cross-referenced against $\mathcal{K}_{\text{prior}}$ and $\mathcal{G}_{\text{evo}}$. The agent assesses whether the observed lesion meets the discriminative features defined in standardized clinical guidelines.
    \item \textbf{Stage 3: Empirical Evidence Alignment.} By analyzing $\mathcal{K}_{\text{hist}}$, the agent evaluates the consistency between the current case and confirmed historical precedents, providing empirical grounding that transcends static model probabilities.
    \item \textbf{Stage 4: Conflict Resolution \& Systematic Synthesis.} In cases when statistical predictions $\mathcal{D}_{\text{pre}}$ conflict with clinical guidelines, the agent prioritizes authoritative standards, identifying if the visual model is influenced by non-diagnostic artifacts.
    \item \textbf{Stage 5: Final Diagnostic Determination.} The synthesized evidence is mapped to a final decision $D^*$, derived from the candidate set $\mathcal{D}_{\text{pre}}$.
\end{itemize}

\subsubsection{The curation of Rare Skin Disease Dataset (RSDD)}
To evaluate the generalization capabilities and diagnostic robustness of SkinGPT-X within the long-tail distribution of dermatology, we curated the RSDD, a novel, high-quality benchmark specifically focused on rare skin conditions. As illustrated in Fig.~\ref{fig:experiments3} a, the construction of RSDD followed a rigorous three-step pipeline:

\begin{itemize}
    \item \textbf{1) Disease Selection and Taxonomy Alignment:} The curation of this dataset was conducted in accordance with the \textit{Rare Disease Diagnosis and Treatment Guidelines} published by the National Health Commission of China\cite{NHC2025RareDiseaseGuidelines}. From this official compendium, 12 dermatological conditions were initially identified. To ensure the integrity of the evaluation and prevent data leakage, we cross-referenced these with our primary training corpus. Overlapping conditions (e.g., Melanoma) were systematically excluded.
    \item \textbf{2) Data Acquisition from Diverse Sources:} To ensure high-fidelity and authoritative clinical representation, we aggregated samples from multiple reputable medical platforms and repositories, including \textit{Wiley Online Library (https://onlinelibrary.wiley.com/)}, \textit{Stamford Skin Center (https://stamfordskin.com/)}, \textit{Wikidoc (https://www.wikidoc.org/)},  \textit{Healthline (https://www.healthline.com/)},  \textit{Healthline (https://www.healthline.com/)},  \textit{ResearchGate (https://www.researchgate.net/)}, and \textit{The National Psoriasis Foundation (https://www.psoriasis.org/)}.
    \item \textbf{3) Final Cohort and Experimental Design:} The final dataset ($n=564$) comprises 8 distinct rare conditions, including \textit{Cutaneous Neuroendocrine Carcinoma} (115), \textit{Generalized Pustular Psoriasis} (110), \textit{Behcet’s Disease} (81), and \textit{Blue Rubber Bleb Nevus Syndrome} (73), \textit{Epithelioid Sarcoma} (46), \textit{Cryopyridine-associated periodic syndrome} (22), \textit{Dermatofibrosarcoma Protuberans} (59), \textit{Gorlin Syndrome} ((58)). To simulate the data-scarce reality of rare disease clinical practice, we implemented a $1:2$ train-test split. This constrained distribution is designed to stress-test the model's few-shot learning efficiency and its ability to generalize from limited clinical presentations compared to general-purpose LLMs.
\end{itemize}
\subsection{Implementation Details}
During the finetuning process of Disease Diagnosis Agent, the max number of epochs is fixed to 30, the warmup epoch is fixed to 10, the learning rate is set to 5e-4 and the batch size is fixed to 128. For other agents in SkinGPT-X, we use Qwen3 series models as the foundation, the maximum token number is fixed to 4096 and the temperature is set to 0.3. To deploy SkinGPT-X entirely locally, a Linux system(e.g. Ubuntu 18.04) is mandatory. For acceleration, we recommend using 8 4090 GPUs with 24GB of memory. SkinGPT-X is developed using Python 3.10, PyTorch 1.10.2 and CUDA 12.5. 
\section{Acknowledgements}

\noindent \textbf{Funding:} This work is supported by The Chinese University of Hong Kong, Shenzhen (CUHK-Shenzhen), under Award No. UDF01004172.

\vspace{\baselineskip}

\noindent \textbf{Competing Interests:} The authors have declared no competing interests. 

\vspace{\baselineskip}

\noindent \textbf{Author Contribution Statements:} Z.C., Y.S. and J.Z. conceived of the presented idea. Z.C., F.W., H.C. and W.D. designed the computational framework and analysed the data. Z.C., Y.X., L.S. and J.Z. conducted the clinical evaluation. J.Z. supervised the findings of this work. Z.C., Y.S., Z.W. and J.Z. took the lead in writing the manuscript and supplementary information. All authors discussed the results and contributed to the final manuscript. 

\vspace{\baselineskip}

\noindent \textbf{Data availability:} The data that support the findings of this study are divided into two groups: shared data and restricted data. Shared data include the HAM10000, DDI, Fitzpatrick-17k, and Dermnet datasets. The HAM10000 dataset is accessible via the Harvard Dataverse at \url{https://doi.org/10.7910/DVN/DBW86T}; the DDI dataset can be accessed through its official project repository at \url{https://ddi-dataset.github.io/}; the Fitzpatrick-17k dataset is available via its GitHub repository at \url{https://github.com/mravuri/fitzpatrick17k}; and the Dermnet dataset can be accessed at \url{https://www.kaggle.com/datasets/shubhamgoel27/dermnet}. Restricted data, including private clinical records and certain proprietary benchmarks, are not publicly available due to patient privacy and institutional data-sharing agreements. The restricted in-house skin disease images used in this study are not publicly available due to restrictions in the data-sharing agreement.

\vspace{\baselineskip}

\noindent \textbf{Code availability:} To promote academic exchanges, under the framework of data and privacy security, the code proposed by SkinGPT-X is publicly available at \url{https://github.com/healme-225040511/Skingpt_X}. In the case of non-commercial use, researchers can sign the license provided in the above link and contact J.Z. or Z.C.
\bibliographystyle{IEEEtran}
\bibliography{main}
\end{document}